\relax
\documentclass[letterpaper]{article} 
\usepackage{aaai18}  
\usepackage{times}  
\usepackage{helvet}  
\usepackage{courier}  
\usepackage{url}  
\usepackage{graphicx}  
\usepackage[utf8]{inputenc} 
\usepackage[T1]{fontenc}    
\usepackage{hyperref}       
\usepackage{url}            
\usepackage{booktabs}       
\usepackage{amsfonts}       
\usepackage{nicefrac}       
\usepackage{microtype}      
\usepackage[disable]{todonotes}
\usepackage{amsmath,amsfonts,amsthm,bm} 
\usepackage{graphicx}
\usepackage{braket}
\usepackage{pgfplots}
\usepackage{tikz}
\usepackage{resizegather}
\usepackage{subcaption}
\usepgfplotslibrary{groupplots}

\usepackage{mathtools}
\usepackage{comment}
\usepackage[backend=bibtex8,style=numeric,doi=false,isbn=false,url=false]{biblatex}
\begin{document}
%
\title{Latent Gaussian Process Regression}
\author{
  Erik Bodin \\
  Department of Computer Science\\
  University of Bristol\\
  Bristol, United Kingdom \\
  \texttt{erik.bodin@bristol.ac.uk} \\
  \And
  Neill D. F. Campbell \\
  Department of Computer Science\\
  University of Bath\\
  Bath, United Kingdom \\
  \texttt{n.campbell@bath.ac.uk} \\
  \And
  Carl Henrik Ek \\
  Department of Computer Science\\
  University of Bristol\\
  Bristol, United Kingdom \\
  \texttt{carlhenrik.ek@bristol.ac.uk} \\
}

\maketitle
\begin{abstract}
 We introduce Latent Gaussian Process Regression which is a latent variable extension allowing modelling of non-stationary multi-modal processes using GPs. The approach is built on extending the input space of a regression problem with a latent variable that is used to modulate the covariance function over the training data. We show how our approach can be used to model multi-modal and non-stationary processes. We exemplify the approach on a set of synthetic data and provide results on real data from motion capture and geostatistics.
\end{abstract}

\section{Introduction}
Gaussian processes (GPs) are probabilistic objects that can be employed as priors to specify distributions over spaces of functions. This provides models with principled uncertainty specification and allows for Bayesian regularization to  balance model complexity with model fit. The flexibility of GPs stems from their non-parametric structure where the characteristics of the prior is fully encapsulated in the choice of covariance and mean function. In all but few cases, the mean function is set to be constant leaving only the covariance function to be chosen. 

Most covariance functions are stationary which means that there is a single structure of variations independent of location in the input space. However, for many types of data, the assumption of a stationary process is not suitable making non-stationary covariances desirable. Creating such covariances often leads to an explosion in the number of parameters, effectively removing the benefit of a non-parametric model. 

An additional challenge with GPs is that they are limited to modelling a single function. Often we have data where, in certain parts of the input space, the data has been generated by several different functions. In such scenarios we desire a model that switches automatically between functions allowing the data to be represented by several different processes.

In this paper we present a unified framework that tackles both these problems. It allows modelling structures, such as non-stationary and multi-modal functions, using GPs without an explosion in the number of parameters. Specifically, we extend \emph{any} covariance function with an additional latent space that encapsulates this structure in a non-parametric manner leading to a single GP with a specific covariance function. 

During inference, we marginalise out these latent variables from the model using the variational approach of \cite{Titsias:2010tb}. This method depends on computing expectations over the covariance function of the GP. This is only analytically tractable for a subset of covariance functions, limiting the applicability of the approach. This motivates the second contribution of this paper. We show that these expectations can be approximated efficiently using Monte Carlo methods, yielding otherwise intractable covariances (such as ours) tractable. It is also beneficial compared to when the expectations need to be analytically tractable as it allows for rapid prototyping by removing challenging and time consuming derivations.

\section{Background}
\begin{figure*}[t]
\centering
\includegraphics[width=0.7\textwidth]{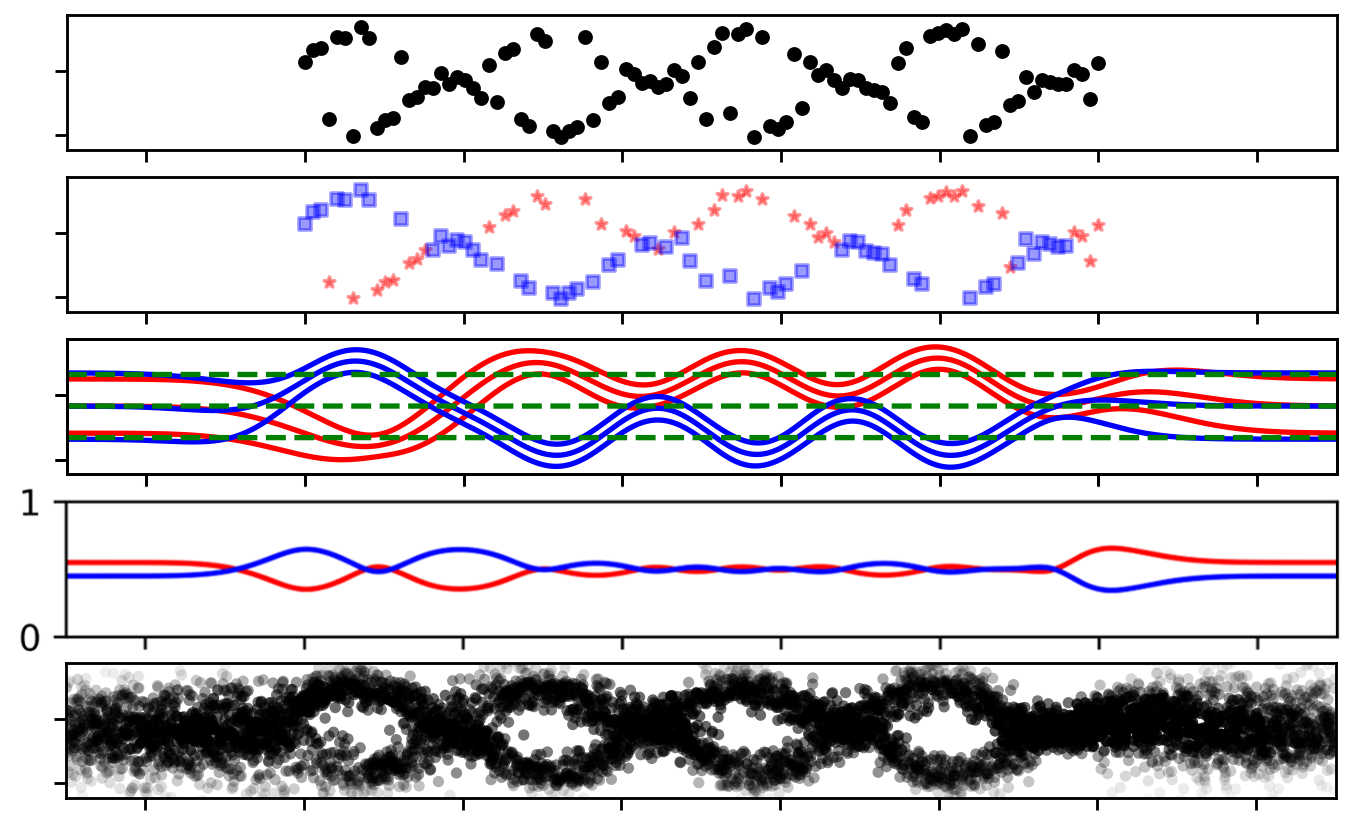}
\caption{\label{fig:antiphase_sine}{\it \small
Factorization. In this experiment the factorizing kernel (\ref{eq:decomposing}) is used to disambiguate two non-uniformly sampled superimposed sinusoids. The factorizing kernel is parameterized by two components, each one a squared exponential kernel with inferred hyper parameters. Upper plot: Synthesized data. Second plot: The component association of each observation. Third plot: Posterior predictions from each component with a standard deviation on each side of the mean. The prediction from a standard GP is shown in green. Fourth plot: Estimated component probabilities (\ref{eq:c_prob}). Bottom plot: Posterior samples.
}}
\end{figure*}
An attractive property of Gaussian processes is that, through very simple means, it is possible to formulate priors that are both interpretable and expressive. Examples are covariance functions, such as the squared exponential, which with a single parameter encode a global smoothness structure. However, for many types of data these global assumptions are not valid. There has been significant interest in how to describe non-stationary covariances allowing for either changing function behaviour, as in \cite{DBLP:conf/icml/AdamsS08}, or for hetroscedastic noise, as in \cite{Lazaro-GredillaT11}. 

In multi-task learning, covariance functions have been defined that are able to model variations between output dimensions such as \cite{Alvarez:2008wy}. As positive-semidefinite kernels are closed under several different operations, there has also been work on how to combine covariances to generate more expressive models. In \cite{2011:Duvenaud}, the authors present an approach where a class of additive covariances is described. In a continuation of this work, the choice of covariance function was formulated as a search problem \cite{DuvLloGroetal13} where a set of base covariances could be combined using additions and multiplications. Using these techniques it is possible to create far more expressive priors while still retaining the benefits of the GP framework.

When moving from modelling stationary to non-stationary covariances, the prior assumption changes from that of a global structure to one of input dependent, local structures. When there are multiple global and/or local trends, these may be modelled by operations on the covariances; for example, as being generated from a sum of globally varying processes in \cite{liutkus2011gaussian} or as a sum of (potentially infinitely many) local experts in \cite{rasmussen2002infinite}. However, all of these models use a single generating mapping from input space to output space. When there are multiple processes generating the data independently, the observations are still forced to be explained in terms of their covariances with respect to all other observations based on their position in the input space. When the underlying data generating processes give rise to multiple modes in the output space, at a single location in the input space, a single-process GP model must resort to explaining the data as noise - failinlg to capture the density of the data and failing to generalize from it. The strength of the GP to model smooth functions is not being utilized in this case as the covariances within each independent \emph{partition} of the observations, produced by their respective underlying data generating process, cancel each other out.

In \cite{lazaro2012overlapping} the authors present an approach to model overlapping GPs, exemplified as a method for e.g. multi-target tracking scenarios and modelling of heteroscedasticity. In their approach, they model data-association of the observations to independent GPs via a latent association matrix. However, since the structure of the latent subspaces created by the association matrix is not explicitly modelled the approach only allows modelling of structures where there is no interdependency between groups. The method we propose in this paper includes this approach as a special case.

The creation of more complicated covariance structures presents a particular challenge during inference.
GPs, in their simplest form, scale cubicly with the data which has lead to a significant amount of work on reducing this computational complexity. In \cite{DBLP:conf/icml/DeisenrothN15} a factorised approach is presented which represents the GP as a product-of-experts that allows for massively distributed computations. However, creating a factorised model such as this relies on making independence assumptions that are not necessarily straight forward or are restrictive on modelling power. 

Another approach is to sparsify the GP and use a smaller set of points, referred to as inducing points, to approximate the full model \cite{DBLP:journals/jmlr/CandelaR05}. This then presents a challenge on how to select the inducing points. In \cite{Titsias:2009vf} a variational approach was presented that learns these inducing points by viewing them as variational parameters specifying the bound. Using the same approach, the authors showed that an extension of the same idea applies to unsupervised learning with GPs \cite{Titsias:2010tb} facilitating approximative integration of the latent variables. The downside of the variational approach presented in \cite{Titsias:2009vf,Titsias:2010tb} is that it requires calculations of expectations over the covariance function. These calculations cumbersome and sometimes intractable which limits scope of applicable covariance functions.

In this paper, we describe a simple extension to any covariance function that allows modelling of non-stationary multi-modal behaviour. Our formulation can model both non-stationary functions, i.e.~when the behaviour of the function is different in different parts of the input domain, and also multi-modal functions where a single input location can be associated with several different outputs. Our approach is based on combining any covariance function with an additional covariance over a latent input space. Using this approach we create a non-parametric model for non-stationary and multi-modal data. We approximately integrate out the latent space using the variational approach in \cite{Titsias:2010tb}. We show empirically that it is possible to approximate the challenging expectations using efficient sampling. This makes our approach applicable to any type of covariance independent of whether or not the expectations can be computed in closed form.

\section{Latent Gaussian Process Regression}
\begin{figure*}[t]
  \centering
  \begin{subfigure}{0.45\linewidth}
    \centering
    \includegraphics[width=\linewidth]{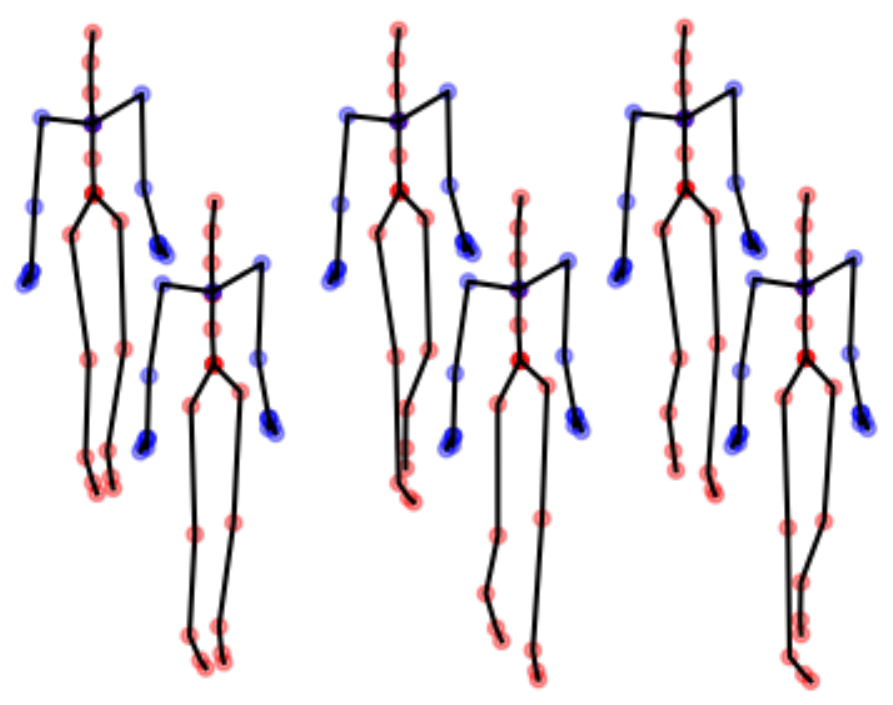}
  \end{subfigure}
  \begin{subfigure}{0.45\linewidth}
    \centering
    \includegraphics[width=\linewidth]{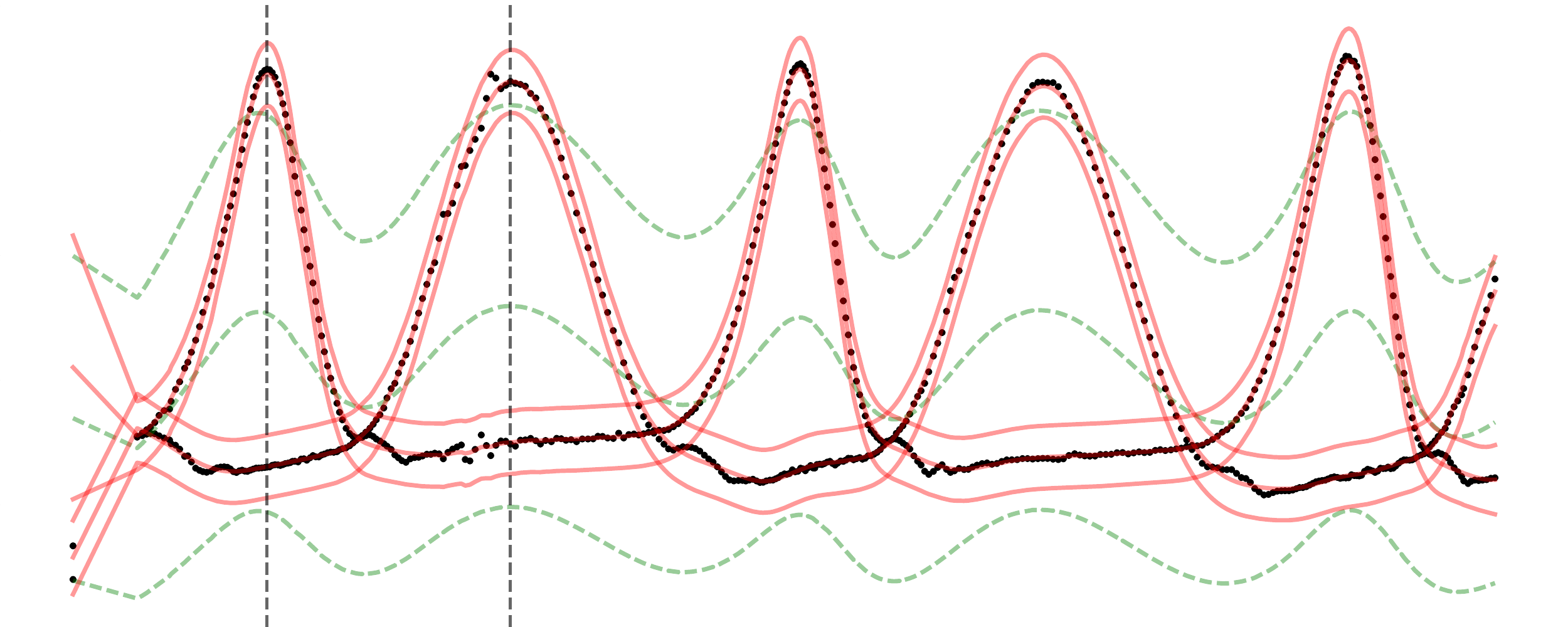}
  \end{subfigure}  
  \caption{\label{fig:motion}Disambiguating motion. In this experiment the factorizing kernel (\ref{eq:decomposing}) is used to disambiguate joint positions in motion data. The character is walking through a room twice; once starting with the left leg and once starting with the right leg. Thus, the joint locations for the legs are ambiguous at each step, causing a 'single-process' GP to predict the mean of the two legs. The presented kernel successfully disambiguates the legs and thus preserves their motions. The training data $\mathbf{X} \in \mathbf{R}^{\mathrm{718 \times 54}}, \mathbf{Y} \in \mathbf{R}^{\mathrm{718 \times 60}}$ is 718 frames of 18 joints (chest, shoulders, arms and hands) as input and 20 joints (legs, spine and head) as output in 3D-space. Note that some of the joints are superimposed for this particular skeleton. The same squared exponential kernel with inferred lengthscale was used to parameterize the factorizing kernel as two kernel components. The upper figure shows the predicted joint locations (in red) at the input joint locations (in blue) for the character at two frames corresponding to consecutive steps using a single-process GP and using the factorizing kernel for the two components, respectively. The lower figure shows the predicted height location for the left heel given a current length-wise location across the room for the left hand, with the single-process GP in green and the two factorized components in red. The black dots in the lower figure corresponds to training data and the vertical lines mark the length-wise location for the two frames in the upper figure. The motion sequence is subject 35 sequence 1 from CMU Graphics Lab Motion Capture Database \cite{gross2001cmu}, with a duplicate of flipped leg motions concatenated with the original.
 }
\end{figure*}  

Consider a set of $N$ input output pairs $\mathcal{D}=\{\mathbf{x}_i,\mathbf{y}_i\}_{i=1}^N$  generated from a non-stationary process $\mathbf{y}_i=g(\mathbf{x}_i)+\epsilon$. We can describe this function as an expansion of processes,
\begin{align}
  \mathbf{y}_i = \sum_j \alpha^{(j)}(\mathbf{x}_i)f^{(j)}(\mathbf{x}_i) + \epsilon
  \label{eq:func}
\end{align}
where $f^{(j)}$ is modelled by a GP. The functions  $\alpha^{(j)}(\mathbf{x}_i)$ modulate the base functions $f^{(j)}$ and encode non-stationary multi-modal behaviour by smoothly segmenting the different functions over the training data. Modelling a sum of functions as a GP is straight forward so the main challenge is how to parametrise $\alpha^{(i)}(\mathbf{x}_i)$. The approach that we take in this paper is to extend the input domain with a latent variable $\mathbf{x}^{(c)}$, such that we have
\begin{align}
  \mathbf{y}_i = \sum_j f^{(j)}(\mathbf{x}_i,\mathbf{x}^{(c)}_i) + \epsilon.
  \label{eq:func2}
\end{align}
Now, rather than directly modulating the output of the function, the latent variable can be used to modulate the covariance function in a non-parametric manner. This allows for modelling of non-functions by differentiating between several outputs at the same input location $\mathbf{x}$ by altering $\mathbf{x}^{(c)}$. This leads to the following latent regression model,
 \begin{align}
 \begin{split}
  \MoveEqLeft p(\mathbf{Y},\mathbf{F},\mathbf{X}^{(c)}|\mathbf{X}) = \\ 
  & p(\mathbf{Y}|\mathbf{F}) \, p(\mathbf{F}|\mathbf{X},\mathbf{X}^{(c)}) \, p(\mathbf{X}^{(c)}),
\end{split}
\end{align}
where $p(\mathbf{F}|\mathbf{X},\mathbf{X}^{(c)})$ is a GP prior over additive functions. By marginalising out the latent variables $\mathbf{X}^{(c)}$ and the GP prior we can recover the standard marginalised likelihood for GP regression,
\begin{align}
\begin{split}
  \MoveEqLeft p(\mathbf{Y}|\mathbf{X}) =\\ 
  & \int p(\mathbf{Y}|\mathbf{F}) \,p(\mathbf{F}|\mathbf{X},\mathbf{X}^{(c)}) \,p(\mathbf{X}^{(c)}) \, \textrm{d}\mathbf{F} \, \textrm{d}\mathbf{X}^{(c)}.
\end{split}
\end{align}
An intuition behind this approach is to think of the marginalisation as a projection, where multiple single-modal processes over the extended input space becomes multi-modal when projected onto the subspace of the original data. We will now describe how we achieve this by using a simple class of covariance functions that we will refer to as juxtaposition kernels.

\begin{figure*}[t]
\begin{minipage}{.5\textwidth}
\centering
\includegraphics[width=\textwidth]{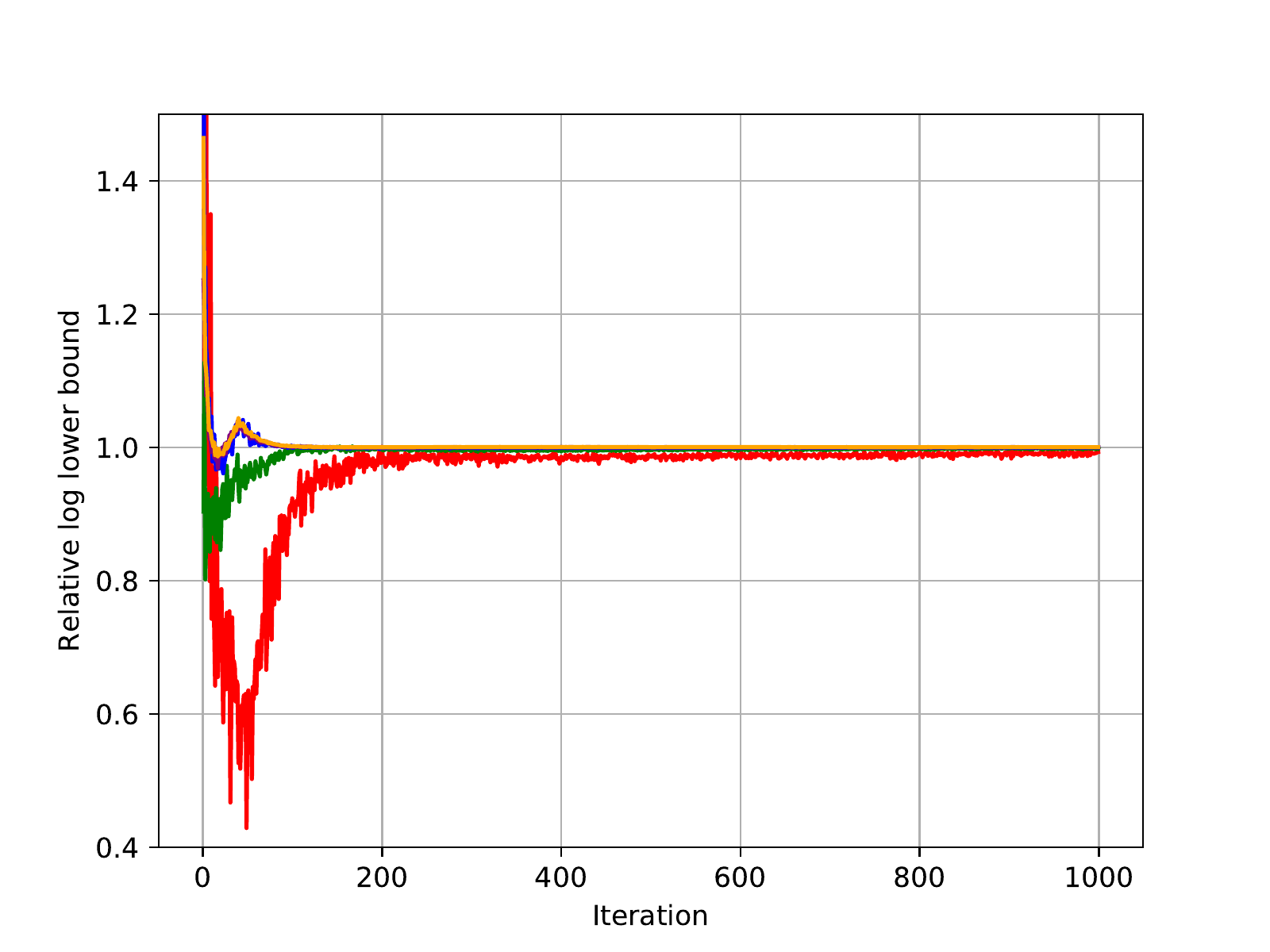}
\end{minipage}%
\begin{minipage}{.5\textwidth}
\centering
\includegraphics[width=\textwidth]{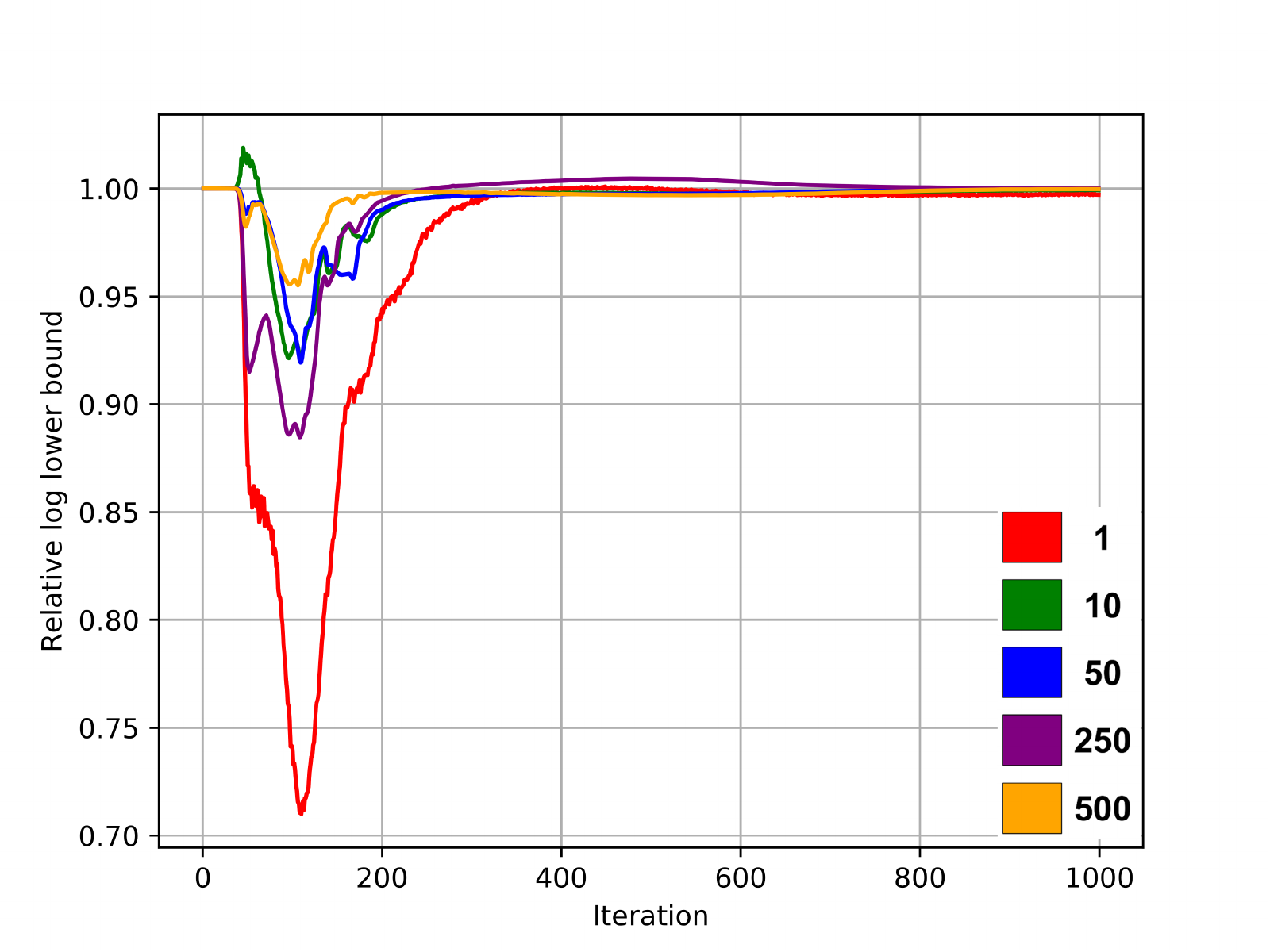}
\end{minipage}
\caption{\label{fig:lowerbound}{\it \small
Comparison of the log lower bounds obtained using Monte Carlo sampling relative to the analytical expectation throughout 1000 iterations for two data sets. Sampling using 1, 10, 50, 250 and 500 samples per iteration are represented by red, green, blue, purple and orange respectively. The exact and the approximated log lower bounds have been normalized with respect to the exact starting and final log lower bound. Left: $\mathbf{x} \in \mathbf{R}^{\mathrm{1}}$, $\mathbf{Y} \in \mathbf{R}^{\mathrm{100 \times 50}}$ is a toy data set comprising 50 draws from a GP with a squared exponential kernel. Right: $\mathbf{x} \in \mathbf{R}^{\mathrm{40}}$, $\mathbf{Y} \in \mathbf{R}^{\mathrm{46 \times 2048}}$ is a data set of font splines used in~\cite{CampbellSIGGRAPH14}.
}}
\end{figure*}

\subsection{Juxtaposition Covariance}
In order to achieve the desired characteristics described above, we will study covariances that are defined as sums of products of different covariance functions evaluated in both the original and the extended input space as,
\begin{equation}
\label{eq:juxta}
k_{\text{juxta}}(\mathbf{x}_i, \mathbf{x}_j) =  \sum_{l=0}^\mathrm{L} w_{l}(\mathbf{x}^{(l)}_i, \mathbf{x}^{(l)}_j) \, k_{l}(\mathbf{x}_i, \mathbf{x}_j),
\end{equation}
%
%
where $w_{l}(\cdot,\cdot)$ is a kernel function over latent inputs $\mathbf{X}^{(l)}$, a subset of $\mathbf{X}^{(c)}$, and $k_{l}(\cdot,\cdot)$ is a kernel function over the observed inputs $\mathbf{X}$. This can be viewed as a GP-prior consisting of a weighted sum of $L$ different kernels $k_{l}(\cdot,\cdot)$ where $w_{l}(\mathbf{x}^{(l)}_i, \mathbf{x}^{(l)}_j)$ describes the influence of $k_{l}(\cdot,\cdot)$ for the input pair $\mathbf{x}_i$ and $\mathbf{x}_j$. In this paper we will evaluate a special case of this kernel that encourages a factorised representation such that the covariance of each pair of observed data points is modelled by either a single kernel function $k_{l}(\cdot,\cdot)$ or considered independent. This is achieved by using a linear kernel over the latent space,
\begin{equation}
	w_{l}(\mathbf{x}_{i}^{(l)}, \mathbf{x}_{j}^{(l)}) = {\mathbf{x}_{i}^{(l)}}^{\mathrm{T}}\mathbf{x}_{j}^{(l)},
\end{equation}
and constraining the latent locations to reside in the corners of a simplex in the unified latent space $\mathbf{X}^{(c)}$ such that $||\mathbf{x}_i^{(c)}||_1 = 1, \forall i$. In this special case, the compound kernel (\ref{eq:juxta}) can be interpreted as a continuous convex association of each observation to a respective component kernel. To achieve this, a transformation is deployed as,
\begin{equation}
\label{eq:transform}
	\varphi(x^{(l)}) = \dfrac{{x^{(l)}}^{\alpha}}{\sum_{l'=0}^\mathrm{L} {x^{(l')}}^{\alpha}},
\end{equation} 
where $\alpha \in \mathbb{R}$ encodes the strength of the discritisation. Since $w_{l}$ is linear and $\mathbf{x}^{(l)}$ is constrained to a simplex it is sufficient for it to be one-dimensional, denoted $x^{(l)}$. To remove effects of initialisation, we use a simple annealing scheme to set $\alpha$ starting with a small value that increases each iteration of the optimisation. The motivation behind this is that with a small value, the associations can easily be altered while it is associated with a significantly higher ``cost'' for large alphas. This yields a \emph{factorising} variant of the juxtaposition kernel, summarized as,
\begin{equation}
\label{eq:decomposing}
	k_{\mathrm{factorising}}(\mathbf{x}_{i}, \mathbf{x}_{j}) =  \sum_{l=0}^\mathrm{L} {\varphi\!\left(x_{i}^{(l)}\right)}\varphi\!\left(x_{j}^{(l)}\right) \, k_{l}(\mathbf{x}_{i}, \mathbf{x}_{j}).
\end{equation}
Using the factorisation above we aim to address the same problem formulation as in \cite{lazaro2012overlapping} however our method is capable of generalising to any structures over the latent space which can be encoded using different latent space priors and is not limited to the linear  kernel as explained above. Further, by removing the transformation Eq.~\ref{eq:transform} we can allow for continous mixtures of processes rather than a discretely factorised.

Predictive inference for novel input locations can be done through the posterior of the model. However, during prediction the latent locations are not known and need to be marginalised from the model,
\begin{gather}
\begin{split}
\label{eq:posterior}
	 \MoveEqLeft p(\mathbf{y}_{*}|\mathbf{x}_{*}, \mathbf{Y}, \mathbf{X}) \approx \\
	& \sum_{l=0}^{\mathrm{L}} \int p(\mathbf{y}_{*}|\mathbf{x}_{*}, \mathbf{Y}, \mathbf{X}, \mathbf{x}_{*}^{(l)})p(\mathbf{x}_{*}^{(l)}|\mathbf{x}_{*}, \mathbf{Y}, \mathbf{X}) \textrm{d}\mathbf{x}^{(l)}_{*}.
\end{split}
\end{gather}
When the latent locations are confined to lie on the corners of a simplex the integral in Eq.~\ref{eq:posterior} reduces to a sum over those corners. This means that the posterior becomes a Gaussian mixture where the distribution over the latent locations $p(\mathbf{x}_{*}^{(l)}|\mathbf{x}_{*}, \mathbf{Y}, \mathbf{X})$ can be interpreted as mixture coefficients. This distribution is not analytically tractable hence we proceed with an assumption. The predictive uncertainty of each component is comparable. Therefore an expression of the relative certainty can be recovered given a coordinate in the input space,
\begin{align}
\begin{split}
  \label{eq:c_prob}
  \rho_{l}(\mathbf{x_{*}}) &= \dfrac{\sum_{l'=0}^{\mathrm{L}} \sigma_{l'}(\mathbf{x_{*}})}{\sigma_{l}(\mathbf{x_{*}})} \\
  \hat{p}(x^{(l)}_{*} &= 1 | \mathbf{x}_{*}, \mathbf{Y}, \mathbf{X}) = \dfrac{\rho_{l}(\mathbf{x}_{*})}{\sum_{l'=0}^{\mathrm{L}} \rho_{l'}(\mathbf{x}_{*})}, \\
\end{split}
\end{align}
where $\sigma_{l}(\mathbf{x}_{*})$ is the square root of the predictive variance of the model for component $l$. This means that outputs can then be generated from the model at any given input location by first sampling a component according to the probabilities in (\ref{eq:c_prob}) and then sampling from the drawn component. Examples of samples and $\hat{p}(x^{(l)}_{*}= 1 | \mathbf{x}_{*}, \mathbf{Y}, \mathbf{X})$ are provided in Figures~\ref{fig:antiphase_sine}, \ref{fig:heteroscedastic_sine} and \ref{fig:s_shape}.

\subsection{Variational Gaussian Process Latent Variable Model}

The GP-LVM framework used throughout this paper, introduced in \cite{damianou2015variational}, deploys \emph{auxiliary inducing variables} as a mean of marginalizing out the input. For our model the input is $\mathbf{X}$ and $\mathbf{X}^{(c)}$, which we jointly denote $\mathbf{X}^{(s)}$. The objective function is specified as a lower bound on the data evidence. In evaluating the lower bound, the following expectations, referred to as \emph{sufficient statistics}, need to be evaluated:
%
%
\begin{align}
\begin{split}
\label{eq:psi_stats}
\xi &= \braket{\,\mathrm{Tr( \mathbf{K}_{\mathrm{ff}} )}\,}_{q(\mathbf{X}^{(s)})} \\
\Psi &= \braket{\,\mathbf{K}_{\mathrm{fu}}\,}_{q(\mathbf{X}^{(s)})} \\
\Phi &= \braket{\,\mathbf{K}_{\mathrm{uf}}\mathbf{K}_{\mathrm{fu}}\,}_{q(\mathbf{X}^{(s)})},
\end{split}
\end{align}
where $q(\mathbf{X}^{(s)}) \sim \mathcal{N}( \bm{\mu}, \mathbf{S})$ and $\bm{\mu}$ and $\mathbf{S}$ are variational parameters.
These expectations are only analytically tractable for \emph{some} convariance kernels, e.g.~the linear and the squared exponential. Thus, the variational framework remains limited to the class of kernels where these expectations are tractable, significantly suppressing its modelling power. Critically, the class of kernels where these expectations are analytically tractable does not include our juxtaposition kernel.

\subsection{Stochastic Approximations of Expectations}

For many kernels, including ours, evaluating these expectations is not tractable; to make progress, we will proceed with a Monte Carlo approach. Besides enabling the variational framework to be used with a vastly larger set of kernels, it is easy to implement, computationally fast and, as we will show, capable of yielding virtually equivalent lower bounds. Below we provide an intuition for why this is. 

Within the Variational GP-LVM framework \cite{Titsias:2010tb,damianou2015variational}, the form of the approximate posterior over the latent space $q(\mathbf{X}^{(s)})$ is selected to be a known parametric distribution (Gaussian) and each observation's input location is parameterized with an egocentric, independent Gaussian for every input dimension. The expectation approximation accuracy over each entry $(i, j)$ in $\mathbf{X}^{(s)}$ is thus independent of dimensionality. In addition the expectations of $\xi$ and $\Phi$ are aggregates over $\mathbf{X}^{(s)}$, which further reduces the approximation error. Further, when used as part of an iterative optimization, each expectation approximation error over the course of the optimization procedure is independent. Thus the approximation error of the expectations results in a `noisy' gradient which is correct on average (cf.~stochastic gradient descent). In summary, the empirical means can be expressed as,
\begin{align}
\begin{split}
\label{eq:approx_psi_stats}
\xi &\approx  \dfrac{1}{T}\sum_{t=0}^{T-1} \mathrm{Tr}\left( \mathbf{K}^{(t)}_{\mathrm{ff}} \right)\\
\Psi &\approx  \dfrac{1}{T}\sum_{t=0}^{T-1} \mathbf{K}^{(t)}_{\mathrm{fu}}\\
\Phi &\approx  \dfrac{1}{T}\sum_{t=0}^{T-1} \mathbf{K}^{(t)}_{\mathrm{uf}} \mathbf{K}^{(t)}_{\mathrm{fu}}
\end{split}
\end{align}
where $\mathbf{K}^{(t)}_{\mathrm{ff}}$ and $\mathbf{K}^{(t)}_{\mathrm{fu}}$ are obtained using $X^{(s)(t)}_{i, j} \sim \mathcal{N}( \mu_{i, j}, S_{i, j})$.

\paragraph{Implementation} We implement the stochastic approximation using the `reparameterization trick' as discussed in \cite{Kingma:2014} and \cite{DBLP:conf/nips/KingmaMRW14} to ensure we obtain low variance estimates for the expectations. The entire architecture is implemented using the Tensorflow framework \cite{Tensorflow}. This allows us to propagate gradients through the sampling procedure as if they were analytically calculated.

\todo[inline,color=purple]{NC: Added the implementation paragraph above - hope it makes sense?}

We will now proceed with the experiments where we provide empirical evidence that the number of samples $T$ can be small enough to be practical to compute while still obtaining accurate enough approximations of the expectations.


\section{Experiments}
Here, we demonstrate the suitability of Monte Carlo methods for approximating the expectations of~(\ref{eq:psi_stats}) as in~(\ref{eq:approx_psi_stats}). We show this with respect to the effect on the resulting lower bound. In all experiments in this section the squared exponential kernel is used to provide a comparison to the analytical expectation for a common case. Throughout the rest of the paper, the juxtaposition kernel \eqref{eq:decomposing} used is \emph{analytically intractable} and we rely entirely on the sampling method. 

\paragraph{Lower Bound} In Fig.~\ref{fig:lowerbound}, the log lower bounds obtained by using the exact analytical expectations are compared to using Monte Carlo sampling. As can be seen, the lower bound obtained using approximations with 10 or more samples per iteration follows the one obtained by the exact analytical expectation closely. Furthermore, even using just one sample to approximate the expectation the lower bound converges to a value close to the analytical. This is in agreement with the findings in \cite{Kingma:2014} and \cite{DBLP:conf/nips/KingmaMRW14}, and makes intuitive sense since the approximation error at every iteration is independent of the error at other iterations and has zero mean; resulting in a `noisy' gradient of the cost function which is correct on average.


\paragraph{Complexity} In terms of computation time, there is an important difference between the sampling method we deploy and the use of analytic expectations. The complexity of the analytic expectation, and its derivatives, can increase greatly, even to the point of making it intractable, to that of directly evaluating the covariance matrix that is needed, regardless, for both methods. 
This issue is magnified particularly in the case of compound kernels.

In the best case, when using the analytical expectation the time complexity remains unchanged. In the worst case, using e.g.~a simple linear kernel, the sampling method would only increase the complexity by a constant factor $T$ (evaluating the covariance matrix $T$ times) assuming that the expectations are the computational bottleneck. 
However, the covariance matrix evaluations can be run in parallel. If the constant $T$ can be kept low without negatively impacting the obtained lower bound, this can lead to substantial speed-ups (depending on kernel and data size). In our specific environment the total computation time of the lower bound using Monte Carlo sampling with $T = 1$ was less than half compared to using the analytical expectation for the squared exponential kernel (and twice with $T = 10$). Throughout the rest of the paper $T = 1$ is used for the analytically intractable presented kernel.


\subsection{Juxtaposition kernel}

\begin{figure}[t]
\centering
\includegraphics[width=0.4\textwidth]{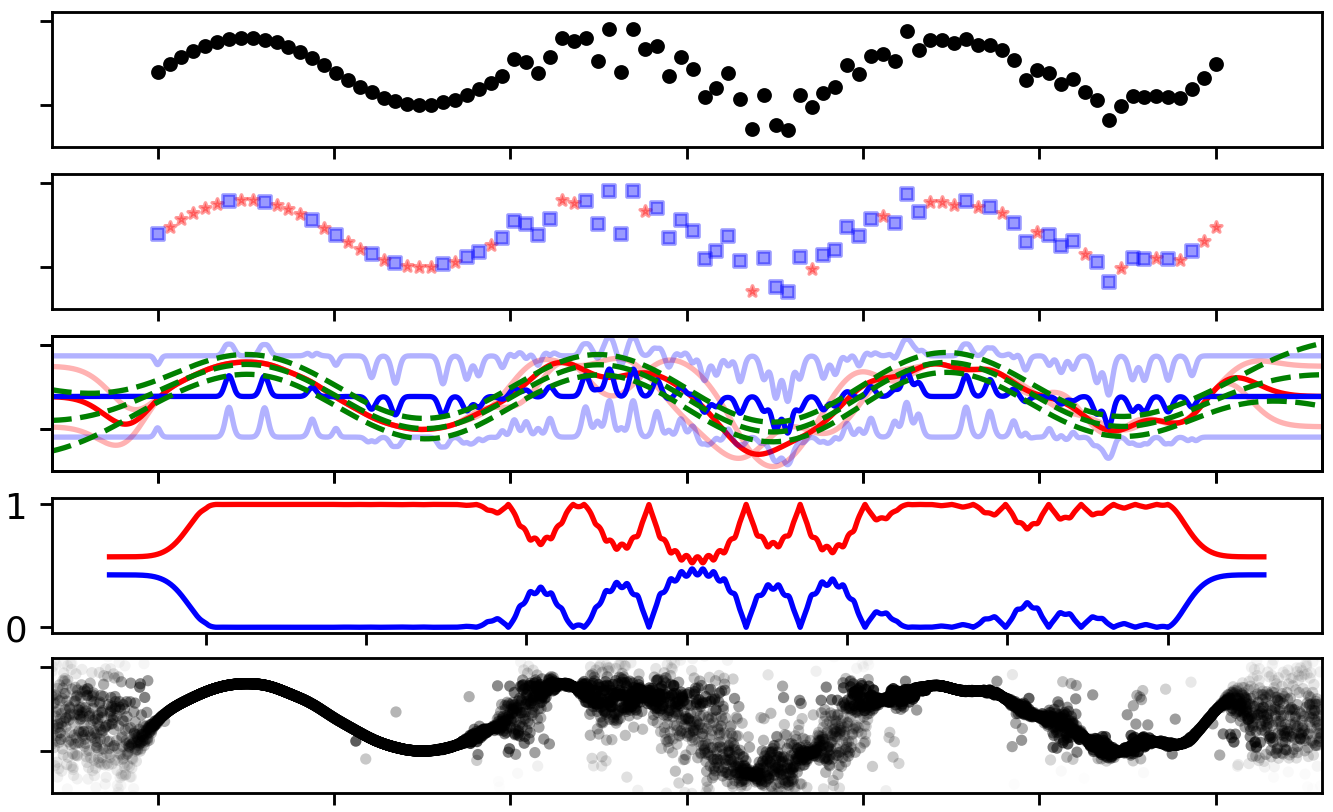}
\caption{\label{fig:heteroscedastic_sine}{\it \small
Heteroscedastic noise via the non-stationary mixing density. The kernel is parameterized by two components; a squared exponential kernel as well as the sum of a squared exponential and a diagonal noise term.  All hyper parameters are inferred. Upper plot: Synthesized data. Second plot: The component association of each observation. Third plot: Posterior predictions from each component with a standard deviation on each side of the mean. The prediction from a standard GP is shown in green. Fourth plot: Estimated component probabilities (\ref{eq:c_prob}). Bottom plot: Posterior samples.
}}
\end{figure}

\paragraph{Multi-Modality} The ability of the model to disambiguate distinct modes in the output space, at the same locations in the input space, is illustrated in Fig.~\ref{fig:antiphase_sine}. Since the two partitions of the observations are conditionally independent they can be explained by different covariance functions and a significantly better data fit can be obtained. Given the regularizing properties of the used variational framework, balance between model complexity and data fit is recovered automatically. 
The data in the example cannot be represented by a Gaussian likelihood satisfactory; this forces a standard (`single-process') GP to explain the data using a high noise variance.  
The result is poorly explained observations and a model with low predictive power. By allowing observations to become conditionally independent of observations close in the input space, via the latent $\mathbf{X}^{(c)}$ space, a more probable and useful explanation of the data is obtained. A real world example of this is illustrated in Fig.~\ref{fig:motion}, where joint positions are disambiguated in motion data.

\begin{figure}[t]
  \centering
  \scalebox{0.9}{\input{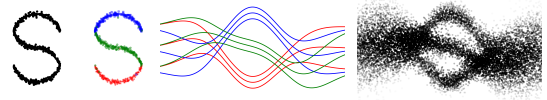}}
    \caption{\label{fig:s_shape}{\it \small
        2-D data as 1-D regression. The kernel is parameterized by three squared exponential kernels. Left plot: Synthesized 2-D. Second plot: The component association of each observation. Third plot: Posterior predictions from each component with a standard deviation on each side of the mean. Right plot: Posterior samples.
    }}
    \label{fig:sshape}
\end{figure}

Posterior predictions of mean and variance are provided by the individual components allowing for prediction in the input space. We compare the results with those of a normal covariance that is forced to explain all of the variations in the signal as noise. Another view-point is to think of our model as a means to cast a multi-modal problem as a regression problem. An extreme case of this is shown in Fig.~\ref{fig:sshape} where the letter {\tt S} is decomposed into three different functions. Importantly, we can predict three different outputs from a single input space, in effect we have decomposed this multimodal regression problem into a regression problem with one free latent variable that differentiates between the modes.

\paragraph{Non-Stationarity} An example of modelling single-modal non-stationarity in the form of heteroscedastic noise is found in Fig.~\ref{fig:heteroscedastic_sine}. Here the factorisation kernel is used as a means of forming a GP prior where individual observations are explained by either a noiseless or noisy smooth function. In this model, the covariance of observations are dependent on the local properties of the data partitioning in the input space (such as relative density) in addition to the hyper parameters governing their respective components covariance function. As a result, the compound covariance function can model  local noise characteristics of the data non-parametrically. By comparison, the standard squared exponential covariance overestimates the variance in the noiseless region while underestimating (being overconfident) in the noisy regions.

\subsection{Jura Geostatistics}

\begin{figure}[t]
\centering
   \begin{subfigure}[b]{0.4\textwidth}
   \includegraphics[width=1\linewidth]{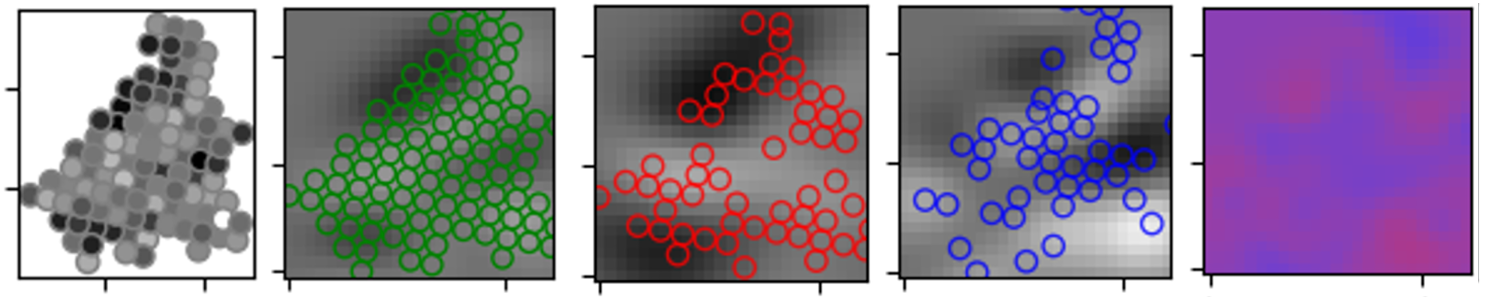}
   \caption{\label{fig:jura}{\it \small Cobalt concentration}}
   \label{fig:Ng1} 
\end{subfigure}\\[5pt]
\begin{subfigure}[b]{0.4\textwidth}
   \includegraphics[width=1\linewidth]{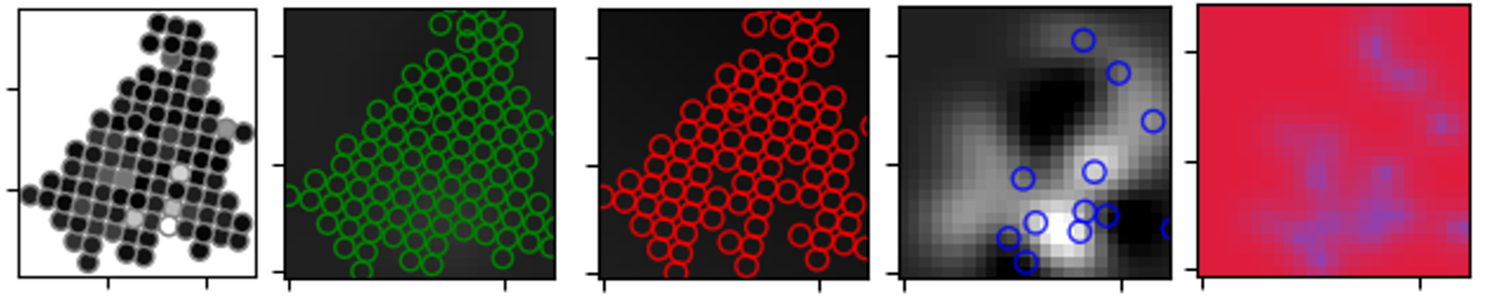}
   \caption{{\it \small Copper concentration}}
   \label{fig:Ng2}
\end{subfigure}

\caption{{\it \small
The factorizing kernel parameterized with two squared exponential kernels to model geostatistics data as a result of a non-stationary probabilistic mixing of two stationary gaussian processes. The data set is from \cite{goovaerts1997geostatistics}. Colored rings illustrate observation component associations. Left column: Element concentrations measured throughout Jura, Switzerland. Second column: Posterior means using a standard GP with a squared exponential kernel (a `single-component' GP). Third and fourth column: Posterior means for the first and second component respectively when used within the factorizing kernel. Right column:  Estimated component probabilities (\ref{eq:c_prob}), where the relative mixture of red and blue illustrate the probability for respective component.
}}
\end{figure}

In Fig.~\ref{fig:jura} the effects of the properties of the presented kernel is illustrated on a real world data set comprised of measured element concentrations throughout the geographical region of Jura, Switzerland \cite{goovaerts1997geostatistics}. The geographical distance between neighbouring data points in the data set are around a kilometer. 
We believe it is fair to assume that the concentrations within the area around each data point are  `sporadically mixed' rather than `homogeneously blended'.
In other words, the soil within an area is not necessarily blended such that individual samples of it contain a representative concentration for the region. An analogy would be trying to locate a suitable location for a gold mine; a single measurement of gold concentration within a square kilometer, even in the most gold rich areas, can result in both a low concentration and `hitting the motherlode'. We model this as that any individual measurement (or sample) within a given area is drawn from any of $\mathrm{L}$ probability distributions. 
The probability for a given  distribution depends on the location, which we model as $\hat{p}(\mathbf{x}^{(l)}_{*}= 1 | \mathbf{x}_{*}, \mathbf{Y}, \mathbf{X})$. For the given data, we make the conservative assumption that $\mathrm{L}=2$ and that independently each of the two generating processes are stationary and smoothly varying; this we model by parameterizing the factorizing kernel using squared exponential kernels. As can be seen in the respective figures, the non-stationary process is not forced to explain the local high concentrations as high levels of global noise but can capture the multi-modal smoothly varying trends of the data set.

\section{Conclusion}
We have presented Latent Gaussian Process Regression, a natural extension to GP regression that allows modelling of non-stationary multi-modal processes using a simple combination of any covariance functions. We show how our method can be used to factorise a signal into several different processes which allows modelling of multi-modal data. We also show how non-stationary single-modal data can be modelled using the same approach. Our approach builds on a latent variable extension of the input domain which is approximatively marginalised out. We provide empirical evidence which highlights that a simple sampling based approach can be used to replace expensive, and sometimes intractable, expectations in the traditional variational formulation of GP-LVMs. In its current form the number of components is a free parameter which we, in later work, hope to directly infer from data.

\printbibliography

\end{document}